\documentclass[10pt, a4paper, logo, twocolumn]{deepmind}
\usepackage{kantlipsum, lipsum}

\usepackage{hyperref}
\usepackage[capitalize,noabbrev]{cleveref}

\newtheorem{definition}{Definition}

\newcommand{\W}{\mathbf{W}}

\renewcommand{\b}{\mathbf{b}}
\renewcommand{\l}{\mathbf{l}}

\newcommand{\E}{\mathbb{E}}

\newcommand{\R}{\mathbb{R}}
\newcommand{\I}{\mathbb{I}}

\newcommand{\dxvec}{\begin{bmatrix}1\\0\\1\end{bmatrix}}
\newcommand{\dyvec}{\begin{bmatrix}0\\1\\-1\end{bmatrix}}

\usepackage[sort&compress, square, numbers]{natbib} 
\graphicspath{{figures/}}

\title{Ray Interference: a Source of Plateaus in Deep Reinforcement Learning}
\correspondingauthor{tom@deepmind.com}

\paperurl{}

\reportnumber{} 

\renewcommand{\today}{2019-04-25}

\author[1,*]{Tom Schaul}
\author[1]{Diana Borsa}
\author[1]{Joseph Modayil}
\author[1]{Razvan Pascanu}

\affil[1]{DeepMind}
\affil[*]{Corresponding author: \texttt{tom@deepmind.com}}

\begin{abstract}
Rather than proposing a new method, this paper investigates an issue present in existing learning algorithms. We study the learning dynamics of reinforcement learning (RL), specifically a characteristic coupling between learning and data generation that arises because RL agents control their future data distribution. In the presence of function approximation, this coupling can lead to a problematic type of \emph{`ray interference'}, characterized by learning dynamics that sequentially traverse a number of performance plateaus, effectively constraining the agent to learn one thing at a time even when learning in parallel is better.
We establish the conditions under which ray interference occurs, show its relation to saddle points and obtain the exact learning dynamics in a restricted setting.
We characterize a number of its properties and discuss possible remedies.
\end{abstract}

\begin{document}
\maketitle
\balance

\section{Introduction}

Deep reinforcement learning (RL) agents have achieved impressive results in recent years, tackling long-standing challenges in board games~\citep{alphago,alphazero}, 
video games~\citep{dqn,OpenAI_dota,alphastarblog} 
and robotics~\citep{openai-shadowhand}.
At the same time, their learning dynamics are notoriously complex. In contrast with supervised learning (SL), these algorithms operate on highly non-stationary data distributions that are \emph{coupled} with the agent's performance: an incompetent agent will not generate much relevant training data.
This paper identifies a problematic, hitherto unnamed, issue with the learning dynamics of RL systems under function approximation (FA).

We focus on the case where the learning objective can be decomposed into \emph{multiple components}, 
\[
J := \sum_{1\leq k\leq K} J_k 
\enspace.
\]
Although not always explicit, complex tasks commonly possess this property.  For example, this property arises when learning about multiple tasks or contexts, when using multiple starting points, in the presence of multiple opponents,  and in domains that contain decisions points with bifurcating dynamics. Sharing knowledge or representations between components can be beneficial in terms of skill reuse or generalization, and sharing seems essential to scale to complex domains. A common mechanism for sharing is a shared function approximator (e.g., a neural network). In general however, the different components do not coordinate and so may compete for resources, resulting in \emph{interference}.

\begin{figure*}[thb]
\begin{center}
\centerline{
\includegraphics[width=0.32\textwidth]{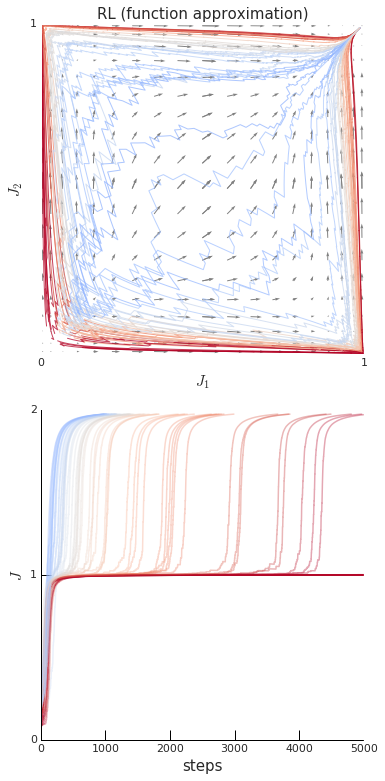}
\hspace{0.02\textwidth}
\includegraphics[width=0.32\textwidth]{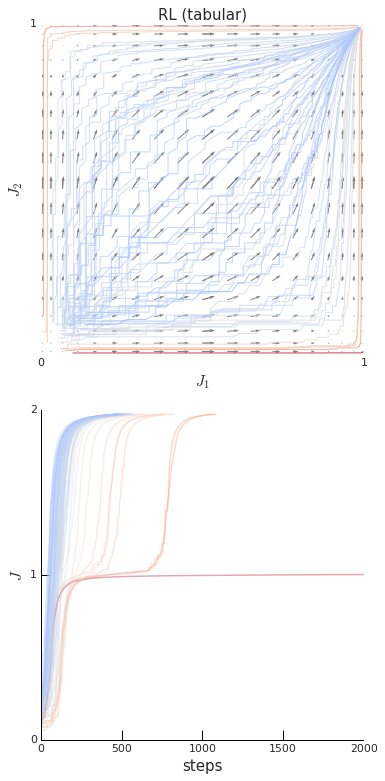}
\hspace{0.02\textwidth}
\includegraphics[width=0.32\textwidth]{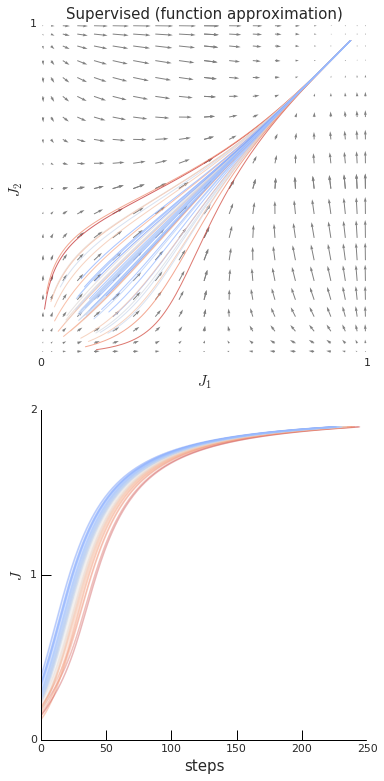}
\vspace{-1em}
}
\caption{Illustration of ray interference in two objective component dimensions $J_1, J_2$. 
{\bf Top row:} Arrows indicate the flow direction of the the learning trajectories.
Each colored line is a (stochastic) sample trajectory, color-coded by performance. 
{\bf Bottom row:} Matching learning curves for these same trajectories.
Note how the trajectories that pass by the saddle points of the dynamics, at $(0,1)$ and $(1,0)$, in warm colors, hit plateaus and learn much slower (note that the scale of the x-axis differs per plot).
Each column has a different setup.
{\bf Left:} RL with FA exhibits ray interference as it has both coupling and interference.
{\bf Middle:} Tabular RL has few plateaus because there is no interference in the dynamics.
{\bf Right:} Supervised learning has no plateaus even with FA interference.
}
\vspace{-1em}
\label{fig:cartoon}
\end{center}
\end{figure*}
The core insight of this paper is that problematic learning dynamics arise when combining two algorithm properties that are relatively innocuous in isolation, the interference caused by the different components and the coupling of the learning signal to the future behaviour of the algorithm.   Combining these properties leads to a phenomenon we call \emph{`ray interference'}\footnote{So called, due to the tentative resemblance of \cref{fig:cartoon} (top, left) to a batoidea (ray fish).}: 
\begin{quote}
\textit{A learning system suffers from plateaus if it has (a) negative interference between the components of its objective, and (b) coupled performance and learning progress.}
\end{quote}
Intuitively, the reason for the problem is that negative interference creates winner-take-all (WTA) regions, where improvement in one component forces learning to stall or regress for the other components of the objective.
Only after such a dominant component is learned (and its gradient vanishes), will the system start learning about the next component.  This is where the coupling of learning and performance has its insidious effect: this new stage of learning is really slow, resulting in a long plateau for overall performance (see \cref{fig:cartoon}).

We believe aspect (a) is common when using neural networks, which are known to exhibit negative interference when sharing representations between arbitrary tasks. On-policy RL exhibits coupling between performance and learning progress (b) because the improving behavior policy generates the future training data. Thus ray interference is likely to appear in the learning dynamics of on-policy deep RL agents.

The remainder of the paper is structured to introduce the concepts and intuitions on a simple example (\cref{sec:bandit}), before extending it to more general cases in \cref{sec:general}. \cref{sec:discussion} discusses prevalence, symptoms, interactions with different algorithmic components, as well as potential remedies.

\section{Minimal, explicit setting}
\label{sec:bandit}
A minimal example that exhibits ray interference is  a $(K\times n)$-bandit problem; a deterministic contextual bandit with $K$ contexts and $n$ discrete actions $a \in \{1, \ldots, n\}$. This problem setting eliminates confounding RL elements of bootstrapping, temporal structure, and stochastic rewards.  It also permits a purely analytic treatment of the learning dynamics.

The bandit reward function is one when the action matches the context and zero otherwise: $r(s_k, a) := \I_{k=a}$ where $\I$ is the indicator function.
The policy is a softmax $\pi(a|s_k) := \frac{\exp( \l_{a,k})}{\sum_{a'}\exp( \l_{a',k})}$, where $\l$ are logits. The mapping from context to logits is parameterised by the trainable weights $\theta$.
The expected performance is the sum across contexts, $J = \sum_k J_k = \sum_k \sum_a r(s_k, a) \pi(a|s_k) = \sum_k \pi(k|s_k)$.

A simple way to train $\theta$ is to follow the policy gradient in REINFORCE~\citep{reinforce}. For a context-action-reward sample, this yields
$\Delta \theta \propto r(s_k, a) \nabla_\theta \log \pi(a|s_k)$.
When samples are generated \emph{on-policy} according to $\pi$, then the expected update in context $s_k$ is
\begin{eqnarray*}
\E[\Delta \theta |s_k] &\propto& 
\sum_{1\leq a\leq n} \pi(a|s_k) \I_{k=a} \nabla_\theta \log \pi(a|s_k) 
\\&=& 
\pi(k|s_k) \nabla_\theta \log \pi(k|s_k) 
\\&=&  
\nabla_\theta  \pi(k|s_k) = \nabla_\theta J_k
\enspace .
\end{eqnarray*}
Interference can arise from function approximation parameters that are shared across contexts. 
To see this, represent the logits as $\l=\W s + \b$, where $\W$ is a $n\times K$ matrix, $\b$ is an $n$-dimensional vector, and  $s \in \mathbb{Z}^{K}_2$ is a one-hot vector representing the current context. Note that each context $s$ uses a different \emph{row} of $\W$, hence no component of $\W$ 
is shared among contexts. However $\b$ is shared among contexts and this is sufficient for interference, defined as follows:
\begin{definition}[Interference]
We quantify the \emph{degree of interference} between two components of $J$ as the cosine similarity between their gradients:
\[
\rho_{k,k'}(\theta) := 
\frac{\langle \nabla J_k(\theta), \nabla J_{k'}(\theta) \rangle}
{\|\nabla J_{k}(\theta)\| \|\nabla J_{k'}(\theta)\|}
\enspace.
\]
 Qualitatively $\rho = 0$ implies no interference, $\rho > 0$ positive transfer, and $\rho<0$ negative transfer (interference).  $\rho$ is bounded between $-1$ and $1$.
\end{definition}

\subsection{Explicit dynamics for a $(2\times 2)$-bandit}

For the two-dimensional case with $K=2$ contexts and $n=2$ arms, we can visualize and express the full  learning dynamics exactly (for expected updates with an infinite batch-size), in the continuous time limit of small step-sizes.
First, we clarify our use of two kinds of derivatives that simplify our presentation. 
The $\nabla$ operator describes a partial derivative and is usually taken with respect to the parameters $\theta$. We omit $\theta$ whenever it is unambiguous, e.g., $\nabla g := \nabla_\theta g(\theta)$. Next, the `overdot' notation for a function $u$ denotes temporal derivatives when following the gradient $\nabla J$ with respect to $\theta$, namely
\[
\dot{u}(\theta) := \lim_{\eta\rightarrow 0} \frac{u\left(\theta + \eta\nabla J\right) - u(\theta)}{\eta}
=\langle \nabla u, \nabla J \rangle.
\]
Let $J_1 = \pi(a=1|s_1)$ and $J_2=\pi(a=2|s_2)$.
For two actions, the softmax can be simplified to a simple sigmoid $\sigma(u)=\frac{1}{1+e^{-u}}$.  By removing redundant parameters, we obtain:
\[
\pi(a=k|s_k) := 
\sigma\left(\theta_1 \I_{k=1,a=1} + \theta_2 \I_{k=2,a=2}
+ \theta_3 (2\I_{a=1}-1)
\right)
\]
 where $\theta := (\W_{1,1}-\W_{1,2}, \W_{2,1}-\W_{2,2}, \b_1-\b_2) \in \R^3$. This yields the following gradients:
\begin{eqnarray*}
\nabla J_1 &=& \nabla_\theta \pi(a=1|s_1) 
 = \nabla_\theta \sigma(\theta_1 + \theta_3) 
\\&=&
\sigma(\theta_1 + \theta_3) (1- \sigma(\theta_1 + \theta_3) ) \frac{d(\theta_1 + \theta_3)}{d\theta}
\\&=&
J_1(1-J_1) \dxvec
\\
\nabla J_2 &=& J_2(1-J_2) \dyvec
\end{eqnarray*}
From these we compute the degree of interference
\begin{eqnarray*}
\rho &:=& \rho_{1,2} = \frac{\langle \nabla J_1, \nabla J_2 \rangle}
{\|\nabla J_1\| \|\nabla J_2\|}
\\&=&
 \frac{J_1(1-J_1) J_2(1-J_2) (-1)}{J_1(1-J_1)\sqrt{2}\cdot J_2(1-J_2)\sqrt{2}} = -\frac{1}{2}
 \enspace,
\end{eqnarray*}
which implies a substantial amount of negative interference at all points in parameter space.

\subsection{Dynamical system}

The learning dynamics follow the full gradient $\nabla J = \nabla J_1 +  \nabla J_2$, and we examine them in the limit of small stepsizes $\eta \rightarrow  0$, in the coordinate system given by $J_1$ and $J_2$. The directional derivatives along $\nabla J$ for the two components $J_1$ and $J_2$ are:
\begin{eqnarray}
\dot{J_1}(\theta) &= &\langle \nabla J_1, \nabla J \rangle
\nonumber\\&=& \|\nabla J_1\|^2 + \langle \nabla J_1, \nabla J_2 \rangle
\nonumber\\&=&  
2J_1^2(1-J_1)^2 - J_1(1-J_1) J_2(1-J_2)
\label{eq:dynamics}
\\
\dot{J_2}(\theta) &=&  2J_2^2(1-J_2)^2 - J_1(1-J_1) J_2(1-J_2)
\label{eq:dynamics2}
\end{eqnarray}
This system of differential equations has fixed points at the four corners, where $(J_1=0,J_2=0)$ is unstable, $(1,1)$ is a stable attractor (the global optimum), and $(0,1)$ and $(1,0)$ are saddle points; see \cref{sec:fpa} for derivations. 
The left of \cref{fig:cartoon} depicts exactly these dynamics.

\subsection{Flat plateaus}

By considering the inflection points where the learning dynamics re-accelerate after a slow-down, we can characterize its plateaus, formally:
\begin{definition}[Plateaus]
We say that the learning dynamics of $J$ have an \emph{$\epsilon$-plateau} at a point $\theta$ if and only if
\begin{eqnarray*}
0 \leq \dot{J}(\theta) \leq \epsilon, &
\ddot{J}(\theta) = 0,&
\dddot{J}(\theta) > 0
\enspace.
\end{eqnarray*}
In other words, $\theta$ is an inflection point of $J$ where the learning curve along $\nabla J$ switches from concave to convex. At $\theta$, the derivative $\epsilon > 0$ characterizes the plateau's {\bf flatness}, characterizing how slow learning is at the slowest (nearby) point.
\end{definition}

In our example, the acceleration is:
\begin{eqnarray}
\label{eq:ddot}
\ddot{J}  &=&
\langle \nabla \dot{J}, \nabla J \rangle 
=(1-J_1-J_2)P_6(J_1,J_2)
\enspace,
\end{eqnarray}
where $P_6$ is a polynomial of max degree 6 in $J_1$ and $J_2$; see \cref{sec:ddot} for details. 
This implies that $\ddot{J}=0$ along the diagonal where $J_2=1-J_1$, and $\ddot{J}$ changes sign there. We have a plateau if the sign-change is from negative to positive, 
see \cref{fig:clines}.
These points lie near the saddle points and are $\epsilon$-plateaus, with their `flatness' given by
\[
\epsilon = \dot{J}|_{J_2=1-J_1} = 2J_1^2(1-J_1)^2
\enspace,
\]
which is vanishingly small near the corners. Under appropriate smoothness constraints on $J$, the existence of an $\epsilon$-plateau slows down learning by $\mathcal{O}(\frac{1}{\epsilon})$ steps compared to a plateau-free baseline; see \cref{fig:eps-badness} for empirical results.

\subsection{Basins of attraction}
Flat plateaus are only a serious concern if the learning trajectories are likely to pass through them. The exact basins of attraction are difficult to characterize, but some parts are simple, namely the regions where one component dominates.
\begin{definition}[Winner-take-all]
We denote the learning dynamics at a point $\theta$ as \emph{winner-take-all} (WTA) if and only if
\[
\dot{J}_k(\theta) > 0
\hspace{2em}\mbox{ and } \hspace{2em}
\forall k' \neq k,  \;\;
\dot{J}_{k'}(\theta) \leq 0 
\enspace ,
\]
that is, following the full gradient $\nabla J$ only increases the $k$\textsuperscript{th} component. 
\end{definition}

\begin{figure}[t]
\begin{center}
\vspace{-1em}
\centerline{
\includegraphics[width=\columnwidth]{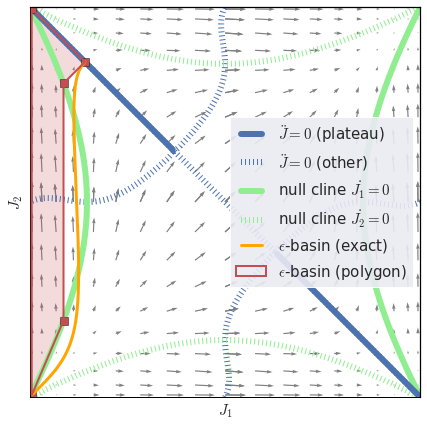}
\vspace{-1em}
}
\caption{
{\bf Bandit learning dynamics:} Geometric intuitions to accompany the derivations. The green hyperbolae show the null clines that enclose the WTA regions. Inflection points are shown in blue, of which the solid lines are plateaus ($\dddot{J}>0$), while the dashed lines are not. The orange path encloses the basin of attraction for a plateau of $\epsilon=0.1$. The red polygon is its lower-bound approximation for which the vertices can be derived explicitly (\cref{sec:lower-bound}).
}
\label{fig:clines}
\end{center}
\end{figure}

The core property of interest for a WTA region is that for every trajectory passing through it, when the trajectory leaves the region, all components of the objective except for the  $k$\textsuperscript{th} will have decreased.
In our example, the null clines describe the WTA regions. They follow the hyperbolae:
\begin{eqnarray*}
\dot{J_1} &=& 0 \Leftrightarrow 2J_1(1-J_1) = J_2(1-J_2)
\nonumber\\
\dot{J_2} &=& 0 \Leftrightarrow J_1(1-J_1) = 2J_2(1-J_2)
\enspace,
\end{eqnarray*}
as shown in \cref{fig:clines}.
Armed with the knowledge of plateau locations and WTA regions, we can establish their \emph{basins of attraction}, see \cref{fig:clines} and \cref{sec:lower-bound}, and thus the likelihood of hitting an $\epsilon$-plateau under any distribution of starting points. For distributions near the origin and uniform across angular directions, it can be shown that the chance of initializing the model in a WTA region is over 50\% (\cref{sec:init-prob}). \cref{fig:eps-basins} shows empirically that for initializations with low overall performance $J(\theta_0) \ll 1$, a large fraction of learning trajectories hit (very) flat plateaus.

\subsection{Contrast example: supervised learning}
\label{sec:supervised}
We can obtain the explicit dynamics for a variant of the $(2\times2)$ setup where the policy is not trained by REINFORCE, but by supervised learning using a cross-entropy loss toward the ground-truth ideal arm, where crucially the performance-learning coupling of on-policy RL is absent:
$
\E[\Delta \theta|s_k] \propto \nabla \log \pi(k | s_k) 
$.
In this case, interference is the same as before ($\rho=- \frac{1}{2}$), but there are no saddle points (the only fixed point is the global optimum at $(1,1)$), nor are there any inflection points that could indicate the presence of a plateau, because $\ddot{J}$ is concave everywhere (see \cref{fig:cartoon}, right, and \cref{sec:sup-detail} for the derivations):
\begin{eqnarray}
\ddot{J}_{\text{sup}} &=& - 2J_1(1-J_1)(1-2J_1 +J_2 )^2 
\nonumber \\&&
- 2J_2(1-J_2)(1-2J_2 +J_1)^2 \leq 0
\enspace.
\label{eq:sup}
\end{eqnarray}

\begin{figure}[h!]
\begin{center}
\vspace{-1em}
\centerline{
\includegraphics[width=\columnwidth]{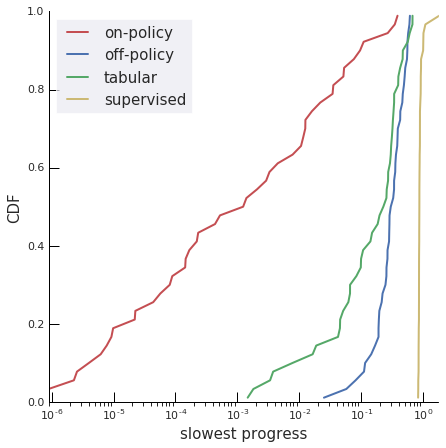}
\vspace{-1em}
}
\caption{
{\bf Likelihood of encountering a flat plateau.}
This plot shows on the likelihood (vertical axis) that the slowest learning progress, $\min |\dot{J}|$, along a trajectory is below some value---when there is a plateau, this is its flatness (horizontal axis). For example, 20\% of on-policy runs (red curve) traverse a very flat plateau with $\epsilon \leq 10^{-5}$.
All these results are empirical quantiles, when starting at low initial performance, $J(\theta_0)=\frac{K}{10}$, and ignoring slow progress near the start or the optimum. There are four settings: ray interference (red) is a consequence of two ingredients, interference and coupling. Multiple ablations eliminate it: interference can be removed by training separate networks or using a tabular representation (green); coupling can be removed by off-policy RL with uniform exploration (blue) or a supervised learning setup as in \cref{sec:supervised} (yellow).
One key contributing factor that impacts whether a trajectory is `lucky' is whether it is initialized near the diagonal ($J_1(\theta_0) \approx J_2(\theta_0)$) or not: the more imbalanced the initial performance, the more likely it is to encounter a slow plateau.
}
\vspace{-1em}
\label{fig:core-diffs}
\end{center}
\end{figure}

\subsection{Summary: Conditions for ray interference}
To summarize, the learning dynamics of a $(2\times 2)$-bandit exhibit ray interference.  For many initializations, the WTA dynamics pull the system into a flat plateau near a saddle point. \cref{fig:cartoon,fig:core-diffs} show this hinges on two conditions.  The negative interference (a) is due to having multiple contexts ($K>1$) and a shared function approximator; this creates the WTA regions that make it likely to hit flat plateaus (left subplots). 

When using a tabular representation instead (i.e., without the action-bias), there are no WTA regions, so the basins of attraction for the plateaus are smaller and do not extend toward the origin, and ray interference vanishes, see \cref{fig:cartoon} (middle subplots). 
On the other hand, the learning dynamics couple performance and learning progress (b) because samples are generated by the current policy. 
Ray interference disappears when this coupling is broken, because without the coupling, the dynamics have no saddle points or plateaus. We see this when a uniform random behavior policy is used, or when the policy is trained directly by supervised learning (\cref{sec:supervised}); see \cref{fig:cartoon} (right subplots).

\section{Generalizations}
\label{sec:general}

In this section, we generalize the intuitions gained on the simple bandit example, and characterize a broader class of problems that exhibit ray interference.

\subsection{Factored objectives}
\label{sec:factored}

There is one class of learning problems that lends itself to such generalization, namely when the component updates are explicitly coupled to performance, and can be written in the following form:
\begin{equation}
\nabla_\theta J_k(\theta) = f_{k}(J_k) v_k(\theta)
\label{eq:ftimesv}
\end{equation}
where $f_k : \R \mapsto \R^+$ is a smooth scalar function mapping to positive numbers that does not depend on the current $\theta$ except via $J_k$, and $v_k : \R^d \mapsto \R^d$ is a gradient vector field.
Furthermore, suppose that each component is bounded: $0 \leq J_k \leq J_k^{\max}$.
When the optimum is reached, there is no further learning, so $f_k(J_k^{\max}) = 0$, but for intermediate points learning is always possible, i.e., $0 < J_k < J_k^{\max} \Rightarrow  f_k(J_k) >0$.

A sufficient condition for a saddle point to exist is that for one of the components there is no learning at its performance minimum, i.e., $\exists k: f_k(0)=0$. The reason is that $f_{k'}(J_{k'}^{\max})=0$, so then $\dot{J}=0$ at any point where all other components are fully learned.

As a first step to assess whether there exist plateaus near the saddle points, we look at the two-dimensional case. Without loss of generality, we pick $f_1(0)=0$ and $J_2^{\max} = 1$.
The saddle point of interest is $(J_1=0, J_2=1)$, so we need to determine the sign of $\ddot{J}$ at the two nearby points $(0, 1-\xi)$ and $(\xi, 1)$, with $0 < \xi \ll 1$. 
Under reasonable smoothness assumptions, a sufficient condition for a plateau to exist between these points is that both $\ddot{J}|_{J_1=0, J_2=1-\xi} <0$ and $\ddot{J}|_{J_1=\xi, J_2=1} > 0$, because $\ddot{J}$ has to cross zero between them.
Under certain assumptions, made explicit in our derivations in \cref{sec:ddot-gen}, we have:
\begin{eqnarray*}
\left.\ddot{J}\right|_{J_1=0} &\approx& 2f_2(J_2)^3 f'_2(J_2) \|v_2\|^4
\\
\left.\ddot{J}\right|_{J_2=1} &\approx& 2f_1(J_1)^3 f'_1(J_1) \|v_1\|^4
\enspace,
\end{eqnarray*}
the sign of which only depends on $f'$. Furthermore, we know that $f'_1(\xi) > 0$ for small $\xi$ because $f_1$ is smooth, $f_1(0)=0$ and $f_1(\xi) >0$, and similarly $f'_2(1-\xi) < 0$ because $f_2(1)=0$ and $f_2(1-\xi) > 0$. In other words, the same condition sufficient to induce a saddle point ($f_1(0)=0$) is also sufficient to induce plateaus nearby.
Note that the approximation here comes from assuming that $\nabla v_k$ is small near the saddle point.

At this point it is worth restating the shape of the $f_k$ in the bandit examples from the previous section: with the REINFORCE objective, we had $f_k(u) =u(1-u)$ and under supervised learning we had $f_k(u) = 1-u$; the extra factor of $u$ in the RL case is what introduced the saddle points, and its source was the (on-policy) data coupling; see \cref{fig:wundt} for an illustration.

Ray interference requires a second ingredient besides the existence of plateaus, namely WTA regions that create the basins of attraction for these plateaus.
For the saddle point at $(J_1=0, J_2=1)$, the WTA region of interest is the one where $J_2$ dominates, i.e., 
\begin{eqnarray*}
 \dot{J_1} \leq 0 &\Leftrightarrow& \langle \nabla J_1, \nabla J \rangle \leq 0
\\ &\Leftrightarrow& f_1(J_1)^2 \|v_1\|^2 + f_1(J_1) f_2(J_2) v_1^{\top}v_2 \leq 0
\\ &\Leftrightarrow& f_1(J_1) \|v_1\|  + \rho  f_2(J_2)  \|v_2\| \leq 0
\enspace.
\end{eqnarray*}
Of course, this can only happen if there is negative interference ($\rho < 0$).  If that is the case however, a WTA region necessarily exists in a strip around $0<\xi\ll 1$, because $f_1$ being smooth means that $f_1(\xi)$ eventually becomes small enough for the negative term to dominate. 
In addition, as \cref{sec:plateau-wta} shows, the sign change in $\ddot{J}$  occurs in the region between the null clines $\dot{J_1}=0$ and $\dot{J_2}=0$, which in turn means that for any plateau, there exist starting points inside the WTA region that lead to it.

\subsection{More than two components}
\label{sec:3dplus}

We have discussed conditions for saddle points to exist for any number of components $K\geq2$. In fact, the number of saddle points grows exponentially with the number of components that satisfy $f_k(0)=0$.
The previous section's arguments that establish the existence of plateaus nearby can be extended to the $K>2$ case as well, but we omit the details here.

Characterizing the WTA regions in higher dimensions is less straight-forward. The simple case is the `fully-interfering' one, where all components compete for the same resources, and they have negative pair-wise interference everywhere ($\forall k\neq k': \rho_{k,k'} <0$): in this case, the previous section's argument can be extended to show that WTA regions must exist near the boundaries.

\begin{figure}[btp]
\begin{center}
\vspace{-1em}
\centerline{
\includegraphics[width=\columnwidth]{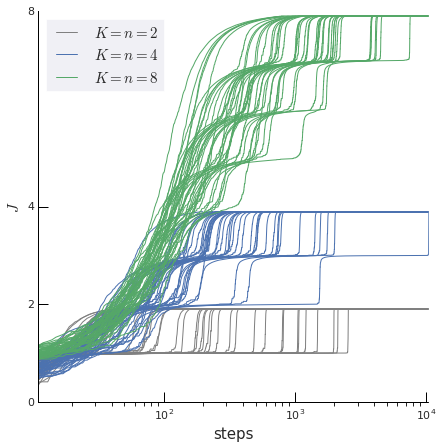}
\vspace{-1em}
}
\caption{
Learning curves when scaling up the problem dimension (jointly $K$ and $n$).
We observe that the $K=8$ runs go through more separate plateaus, and each plateau takes exponentially longer to overcome than the previous one (the horizontal axis is log-scale).
}
\label{fig:beyond2d}
\end{center}
\end{figure}

However, WTA is an unnecessarily strong criterion for pulling trajectories toward plateaus for ray interference: we have seen in \cref{fig:clines} that the basins of attraction extend beyond the WTA region (compare green and orange), especially in the low-performance regime. For example consider three components A, B and C, where A learns first and suppresses B (as before). 
Now during this stage, C might behave in different ways. It could learn only partially, converge in parallel with A or be suppressed as B. When moving to stage two, once A is learned, if C has not fully converged, \emph{ray interference} dynamics can appear between B and C. 
Note that a critical quantity is the performance of C after stage one; this is not a trivial one to make formal statements about, so we rely on an empirical study.
\cref{fig:beyond2d} shows that the number of plateaus grows with $K$ in fully-interfering scenarios. 
For $K=8$, we observe that typically a first plateau is hit after a few components have been learned ($J\approx3$), indicating that the initialization was not in a WTA region. But after that, most learning curves look like step functions that learn one of the remaining components at a time, with plateaus in-between these stages.

A more surprising secondary phenomenon is that the plateaus seem to get exponentially worse in each stage (note the log-scale). We propose two interpretations.
First, consider the two last components to be learned. They have not dominated learning for $K-2$ stages, all the while being suppressed by interference, and thus their performance level is very low when the last stage starts, much lower than at initialization. And as \cref{fig:eps-basins} shows, a low initial performance dramatically affects the chance that the dynamics go through a very flat plateau. 
Second, the length of the plateau may come from the interference of the $K$\textsuperscript{th} task with all previously learned $K-1$ tasks. Basically, when starting to learn the $K$\textsuperscript{th} task, a first step that improves it can negatively interfere with the first $K-1$ tasks. In a second step, these previous tasks may dominate the update and move the parameters such that they recover their performance. 
Thus the only changes preserved from these two tug-of-war steps are those in the \emph{null-space} of the first $K-1$ tasks. Learning to use only that restricted capacity for task $K$ takes time, especially in the presence of noise, and could thus explain that the length the plateaus grows with $K$.

\subsection{From bandits to RL}
While the bandit setting has helped ease the exposition, our aim is to gain understanding in the more general (deep) RL case. In this section, we argue that there are \emph{some} RL settings that are likely to be affected by ray interference. First, there are many cases where the single scalar reward objective can be seen as a composite of many $J_k$: the simple analogue to the contextual bandit are domains with multiple rooms, levels or opponents (e.g., Atari games like Montezuma's Revenge), and a competent policy needs to be good in/against many of these. The additive assumption $J=\sum J_k$ may not be a perfect fit, but can be a good first-order approximation in many cases.
More generally, when rewards are very sparse, decompositions that split trajectories near reward events is a common approximation in hierarchical RL~\citep{lane2001toward,moore1999multi}.
Other plausible decompositions exist in undiscounted domains where each reward can be `collected' just once, in which case each such reward can be viewed as a distinct $J_k$, and the overall performance depends on how many of them the policy can collect, ignoring the paths and ordering.
It is important to note that a decomposition does not have to be explicit, semantic or clearly separable: in some domains, competent behavior may be well explained by a combination of implicit skills---and if the learning of such skills is subject to interference, then ray interference may be an issue.

Another way to explain RL dynamics from our bandit investigation is to consider that each arm is a temporally abstract \emph{option}~\citep{options}, with `contexts' referring to parts of state space where different options are optimal. This connection makes an earlier assumption less artificial: we assumed low initial performance in \cref{fig:core-diffs}, which is artificial in a 2-armed bandit, but  plausible if there is an arm for each possible option.

In RL, interference can also arise in other ways than the competition in policy space we observed for bandits.  There can be competition around what to memorize, how to represent state, which options to refine, and where to improve value accuracy. These are commonly conflated in the dynamics of a shared function approximator, which is why we consider ray interference to apply to \emph{deep} RL in particular.
On the other hand, the potential for interference is also paired with the potential for positive transfer ($\rho >0$), and a number of existing techniques try to exploit this, for example by learning about auxiliary tasks~\citep{unreal}.

Coupling can also be stronger in the RL case than for the bandit: while we considered a uniform distribution over contexts, it is more realistic to assume that the RL agent will visit some parts of state space far more often than others.  It is likely to favour those where it is learning, or seeing reward already (amplifying the rich-get-richer effect). To make this more concrete, assume that the performance  $J_k$ sufficiently determines the data distribution the agent encounters, such that its effect on learning about $J_k$ in on-policy RL can be summarized by the scalar function $f_k(J_k)$ (see \cref{sec:factored}), at least in the low-performance regime. 
For the types of RL domains discussed above it is likely that $f_k(0)\approx 0$, i.e., that very little can be learned from only the data produced by an incompetent policy---thereby inducing ray interference.

We have already alluded to the impact of on-policy versus off-policy learning: the latter is generally considered to lead to a number of difficulties~\citep{suttonbook}---however, it can also reduce the coupling in the learning dynamics; see for example \cref{fig:wundt} for an illustration of how $f_k$ changes when mixing the soft-max policy with 10\% of random actions: crucially it  no longer satisfies $f_k(0)=0$.
This perspective that off-policy learning can induce better learning dynamics in some settings, took some of the authors by surprise.

\subsection{Beyond RL}

A related phenomenon to the one described here was previously reported for supervised learning by Saxe et al. \citep{saxe2013exact}, for the particular case of \emph{deep linear} models. This setting makes it easy to analytically express the learning dynamics of the system, unlike traditional neural networks. Assuming a single hidden layer deep linear model, and following the derivation in \citep{saxe2013exact}, under the full batch dynamics, the continuous time update rule is given by: 
\begin{eqnarray*}
    \dot{W_1} &\propto & W_2^{\top} (\Sigma_{xy} - W_2 W_1\Sigma_{xx})
    \\
    \dot{W_2} &\propto & (\Sigma_{xy} - W_2 W_1\Sigma_{xx}) W_1^{\top} \enspace,
\end{eqnarray*}
where $W_1$ and $W_2$ are the weights on the first and second layer respectively (the deep linear model does not have biases), $\Sigma_{xy}$ is the correlation matrix between the input and target, and $\Sigma_{xx}$ is the input correlation matrix.
Using the singular value decomposition of $\Sigma_{xy}$ and assuming $\Sigma_{xx}=\I$, they study the dynamics for each mode, and find that the learning dynamics lead to multiple plateaus, where each plateau corresponds to learning of a different mode. Modes are learned sequentially, starting with the one corresponding to the largest singular value (see the original paper for full details). Learning each mode of the $\Sigma_{xy}$ has its analogue in the different objective components $J_k$ in our notation. The learning curves showed in \citep{saxe2013exact} resemble those observed in  \cref{fig:beyond2d,fig:cartoon}, and their Equation (5) describing the per-mode dynamics has a similar structure to our Equations \ref{eq:dynamics} and \ref{eq:dynamics2}.

Our intuition on how this similarity comes about is speculative.
One could view the hidden representation as the input of the top part of the model ($W_2$). Now from the perspective of that part, the input distribution is \emph{non-stationary} because the hidden features (defined by $W_1$) change.
Moreover, this non-stationarity is coupled to what has been learned so far, because the error is propagated through the hidden features into $W_1$. 
If the system is initialized such that the hidden units are correlated, then learning the different modes leads to a competition over the same representation or resources.
The gradient is initially dominated by the mode with the largest singular value and therefore the changes to the hidden representation correspond features needed to learn this mode only. 
Once the loss for the dominant mode converges, symmetry-breaking can happen, and some of the hidden features specialize to represent the second mode. This transition is visible as a plateau in the learning curve.

While this interpretation highlights the similarities with the RL case via coupling and competition of resources, we want to be careful to highlight that both of these aspects work differently here.  The coupling does not have the property that low performance also slows down learning (\cref{sec:factored}).  It is not clear whether the modes exhibit negative interference where learning about one mode leads to undoing progress on another one, it could be more akin to the magnitude of the noise of the larger mode obfuscating the signal on how to improve the smaller one.

Their proposed solution is an initialization scheme that ensures all variations of the data are preserved when going through the hierarchy, in line with previous initialization schemes \citep{xavier-init,lecun98-efficient}, which leads to symmetry breaking and reduces negative interference during learning. 
Unfortunately, as this solution requires access to the entire dataset, it does not have a direct analogue in RL where the relevant states and possible rewards are not available at initialization.

\paragraph{Multi-task versus continual learning} 
\label{sec:multi-task}
Our investigation has natural connections to the field of \emph{multi-task} learning (be it for supervised or RL tasks \citep{ruder2017overview,oh2017zero}), namely by considering that the multitask objective is additive over one $J_k$ per task.
It is not uncommon to observe \emph{task dominance} in this scenario (learning one task at a time~\citep{guo2018dynamic}), and our analysis suggests possible reasons why tasks are sometimes learned sequentially despite the setup presenting them to the learning system all at once. On the other hand, we know that deep learning struggles with fully sequential settings, as in \emph{continual learning} or \emph{life-long learning} \citep{ring1994continual,barbados,silver2013lifelong}, one of the reasons being that the neural network's capacity can be exhausted prematurely (saturated units), resulting in an agent that can never reach its full potential. So this raises the questions why current multi-task techniques appear to be so much more effective than continual learning, if  they implicitly produce sequential learning?
One hypothesis is that the potential tug-of-war dynamics that happen when moving from one component to another are akin to rehearsal methods for continual learning, help split the representation, and allowing room for learning the features required by the next component.
Two other candidates could be that the implicit sequencing produces better task orderings and timings than external ones, or the notion that what is sequenced are not tasks themselves but skills that are useful across multiple tasks. But primarily, we profess our ignorance here, and hope that future work will elucidate this issue, and lead to significant improvements in continual learning along the way.

\section{Discussion}
\label{sec:discussion}

\paragraph{How prevalent is it?}
Ray interference does not require an explicit multi-task setting to appear. A given single objective might be internally composed of subtasks, some of which have negative interference. 
We hypothesize, for example, that performance plateaus observed in Atari~\citep[e.g.][]{dqn,rainbow} might be due to learning multiple interfering skills (such as picking up pellets, avoiding ghosts and eating ghosts in Ms PacMan). 
Conversely, some of the explicit multi-task RL setups appear not to suffer from visible plateaus~\citep[e.g.][]{impala}. 
There is a long list of reasons for why this could be, from the task not having interfering subtasks, positive transfer outweighing negative interference, the particular architecture used, or population based training~\citep{pbt} hiding the plateaus through reliance on other members of the population.
Note that the lack of plateaus does not exclude the sequential learning of the tasks.
Finally, ray interference might not be restricted to RL settings. Similar behaviour has been observed for deep (linear) models~\citep{saxe2013exact}, though we leave developing the relationship between these phenomena as future work.

\paragraph{How to detect it?} 
Ray interference is straight-forward to detect if the components $J_k$ are known (and appropriate), by simply monitoring whether progress stalls on some components while others learn, and then picks up later. It can be verified by training a separate network for each component from the same (fixed) data.
In the more general case, where only the overall objective $J$ is known, a first symptom to watch out for is plateaus in the learning curve of \emph{individual runs}, as plateaus tend to be averaged out when curves aggregated across many runs. 
Once plateaus have been observed, there are two types of control experiments: interference can be reduced by changing the function approximator (capacity or structure), or coupling can be reduced by fixing the data distribution or learning more off-policy. If the plateaus dissipate under these control experiments, they were likely due to ray interference.

\paragraph{What makes it worse?}
For simplicity, we have examined only one type of coupling, via the data generated from the current policy, but there can be other sources. When contexts/tasks are not sampled uniformly but shaped into \emph{curricula} based on recent learning progress~\citep{forestier2016modular,act}, this amplifies the winner-take-all dynamics. 
Also, using temporal-difference methods that \emph{bootstrap} from value estimates~\citep{suttonbook} may introduce a form of coupling where the values improve faster in regions of state space that already have accurate and consistent bootstrap targets.
A form of coupling that operates on the population level is connected to selective pressure~\cite{pbt}: the population member that initially learns fastest can come to dominate the population and reduce \emph{diversity}---favoring one-trick ponies in a multi-player setup, for example.

\paragraph{What makes it better?}
There are essentially three approaches: reduce interference, reduce coupling, or tackle ray interfere head-on.
Assuming knowledge of the components of the objective, the multi-task literature offers a plethora of approaches to avoid negative interference, from modular or multi-head architectures to gating or attention mechanisms \citep{fernando2017pathnet,rusu2015policy,stollenga2014deep,chung2014empirical,shazeer2017outrageously,saxe-switching}.
Additionally, there are methods that prevent the interference directly at the gradient level~\citep{du2018adapting,yin2017grad}, normalize the scales of the losses~\citep{popart}, or explicitly preserve capacity for late-learned subtasks~\citep{ewc}. 
It is plausible that following the \emph{natural} gradient~\citep{amari1998natural,peters2005natural} helps as well, see for example \citep{polytope} (their figures 9b and 11b) for preliminary evidence.
When the components are not explicit, a viable approach is to use population-based methods that encourage diversity, and exploit the fact that different members will learn about different implicit components; and that knowledge can be combined~\citep[e.g., using a distillation-based cross-over operator,][]{gangwani2017genetic}.
A possibly simpler approach is to rely on innovations in deep learning itself: it is plausible that deep networks with ReLU non-linearities and appropriate initialization schemes~\citep{he-init} implicitly allow units to specialize.
Note also that the interference can be \emph{positive}, learning one component helps on others (e.g., via refined features).
Coupling can be reduced by introducing elements of off-policy learning 
that dilutes the coupled data distribution with exploratory experience (or experience from other agents), rebalancing the data distribution with a suitable form of prioritized replay~\citep{schaul2015prioritized} or fitness sharing~\citep{sareni1998fitness}, or by reward shaping that makes the learning signal less sparse~\citep{rudder}.
A generic type of decoupling solution (when components are explicit) is to train separate networks per component, and \emph{distill} them into a single one~\citep{lopez2015unifying,rusu2015policy}.
Head-on approaches to alleviate ray interference could draw from the growing body of continual learning techniques~\citep{ring1994continual,barbados,silver2013lifelong,riemer2018learning,lopez2017gradient}.

\section{Conclusion}
This paper studied ray interference, an issue that can stall progress in reinforcement learning systems.  It is a combination of harms that arise from (a) conflicting feedback to a shared representation from multiple objectives, and (b) changing the data distribution during policy improvement. These harms are much worse when combined, as they cause learning progress to stall, because the expected learning update drags the learning system towards plateaus in gradient space that require a long time to escape.  As such, ray interference is not restricted to deep RL (a bias unit weight shared across different actions in a linear model suffices), but rather it shows how harmful forms of interference, similar to those studied in deep learning, can arise naturally within reinforcement learning. 
This initial investigation stops short of providing a full remedy, but it sheds light onto these dynamics, improves understanding, teases out some of the key factors, and hints at possible directions for solution methods.

Zooming out from the overall tone of the paper, we want to highlight that plateaus are not omnipresent in deep RL, even in complex domains.  
Their absence might be due to the many commonly used practical innovations that have been proposed for stability or performance reasons. As they affect the learning dynamics, they could indeed alleviate ray interference as a secondary effect. It may therefore be worth revisiting some of these methods, from a perspective that sheds light on their relation to phenomena like ray interference.

\bibliography{bib}

\begin{thebibliography}{10}

\bibitem{amari1998natural}
S.-I. Amari.
\newblock Natural gradient works efficiently in learning.
\newblock {\em Neural computation}, 10(2):251--276, 1998.

\bibitem{rudder}
J.~A. Arjona-Medina, M.~Gillhofer, M.~Widrich, T.~Unterthiner, J.~Brandstetter,
  and S.~Hochreiter.
\newblock {RUDDER}: Return decomposition for delayed rewards.
\newblock {\em arXiv preprint arXiv:1806.07857}, 2018.

\bibitem{chung2014empirical}
J.~Chung, C.~Gulcehre, K.~Cho, and Y.~Bengio.
\newblock Empirical evaluation of gated recurrent neural networks on sequence
  modeling.
\newblock {\em arXiv preprint arXiv:1412.3555}, 2014.

\bibitem{polytope}
R.~Dadashi, A.~A. Ta{\"{\i}}ga, N.~L. Roux, D.~Schuurmans, and M.~G. Bellemare.
\newblock The value function polytope in reinforcement learning.
\newblock {\em CoRR}, abs/1901.11524, 2019.

\bibitem{du2018adapting}
Y.~Du, W.~M. Czarnecki, S.~M. Jayakumar, R.~Pascanu, and B.~Lakshminarayanan.
\newblock Adapting auxiliary losses using gradient similarity.
\newblock {\em arXiv preprint arXiv:1812.02224}, 2018.

\bibitem{impala}
L.~Espeholt, H.~Soyer, R.~Munos, K.~Simonyan, V.~Mnih, T.~Ward, Y.~Doron,
  V.~Firoiu, T.~Harley, I.~Dunning, et~al.
\newblock {IMPALA}: Scalable distributed {Deep-RL} with importance weighted
  actor-learner architectures.
\newblock {\em arXiv preprint arXiv:1802.01561}, 2018.

\bibitem{fernando2017pathnet}
C.~Fernando, D.~Banarse, C.~Blundell, Y.~Zwols, D.~Ha, A.~A. Rusu, A.~Pritzel,
  and D.~Wierstra.
\newblock Pathnet: Evolution channels gradient descent in super neural
  networks.
\newblock {\em arXiv preprint arXiv:1701.08734}, 2017.

\bibitem{forestier2016modular}
S.~Forestier and P.-Y. Oudeyer.
\newblock Modular active curiosity-driven discovery of tool use.
\newblock In {\em Intelligent Robots and Systems (IROS), 2016 IEEE/RSJ
  International Conference on}, pages 3965--3972. IEEE, 2016.

\bibitem{gangwani2017genetic}
T.~Gangwani and J.~Peng.
\newblock Genetic policy optimization.
\newblock {\em arXiv preprint arXiv:1711.01012}, 2017.

\bibitem{xavier-init}
X.~Glorot and Y.~Bengio.
\newblock Understanding the difficulty of training deep feedforward neural
  networks.
\newblock In {\em In Proceedings of the International Conference on Artificial
  Intelligence and Statistics (AISTATS’10). Society for Artificial
  Intelligence and Statistics}, 2010.

\bibitem{act}
A.~Graves, M.~G. Bellemare, J.~Menick, R.~Munos, and K.~Kavukcuoglu.
\newblock Automated curriculum learning for neural networks.
\newblock {\em arXiv preprint arXiv:1704.03003}, 2017.

\bibitem{guo2018dynamic}
M.~Guo, A.~Haque, D.-A. Huang, S.~Yeung, and L.~Fei-Fei.
\newblock Dynamic task prioritization for multitask learning.
\newblock In {\em Proceedings of the European Conference on Computer Vision
  (ECCV)}, pages 270--287, 2018.

\bibitem{he-init}
K.~He, X.~Zhang, S.~Ren, and J.~Sun.
\newblock Delving deep into rectifiers: Surpassing human-level performance on
  imagenet classification.
\newblock {\em CoRR}, abs/1502.01852, 2015.

\bibitem{rainbow}
M.~Hessel, J.~Modayil, H.~Van~Hasselt, T.~Schaul, G.~Ostrovski, W.~Dabney,
  D.~Horgan, B.~Piot, M.~Azar, and D.~Silver.
\newblock Rainbow: Combining improvements in deep reinforcement learning.
\newblock In {\em Thirty-Second AAAI Conference on Artificial Intelligence},
  2018.

\bibitem{popart}
M.~Hessel, H.~Soyer, L.~Espeholt, W.~Czarnecki, S.~Schmitt, and H.~van Hasselt.
\newblock Multi-task deep reinforcement learning with popart.
\newblock {\em arXiv preprint arXiv:1809.04474}, 2018.

\bibitem{pbt}
M.~Jaderberg, V.~Dalibard, S.~Osindero, W.~M. Czarnecki, J.~Donahue, A.~Razavi,
  O.~Vinyals, T.~Green, I.~Dunning, K.~Simonyan, et~al.
\newblock Population based training of neural networks.
\newblock {\em arXiv preprint arXiv:1711.09846}, 2017.

\bibitem{unreal}
M.~Jaderberg, V.~Mnih, W.~M. Czarnecki, T.~Schaul, J.~Z. Leibo, D.~Silver, and
  K.~Kavukcuoglu.
\newblock Reinforcement learning with unsupervised auxiliary tasks.
\newblock {\em arXiv preprint arXiv:1611.05397}, 2016.

\bibitem{ewc}
J.~Kirkpatrick, R.~Pascanu, N.~Rabinowitz, J.~Veness, G.~Desjardins, A.~A.
  Rusu, K.~Milan, J.~Quan, T.~Ramalho, A.~Grabska-Barwinska, et~al.
\newblock Overcoming catastrophic forgetting in neural networks.
\newblock {\em Proceedings of the national academy of sciences},
  114(13):3521--3526, 2017.

\bibitem{lane2001toward}
T.~Lane and L.~P. Kaelbling.
\newblock Toward hierarchical decomposition for planning in uncertain
  environments.
\newblock In {\em Proceedings of the 2001 IJCAI workshop on planning under
  uncertainty and incomplete information}, pages 1--7, 2001.

\bibitem{lecun98-efficient}
Y.~LeCun, L.~Bottou, G.~B. Orr, and K.-R. Müller.
\newblock Efficient backprop.
\newblock In G.~Montavon, G.~B. Orr, and K.-R. Müller, editors, {\em Neural
  Networks: Tricks of the Trade}, volume 7700 of {\em Lecture Notes in Computer
  Science}, pages 9--48. Springer, 1998.

\bibitem{lopez2015unifying}
D.~Lopez-Paz, L.~Bottou, B.~Sch{\"o}lkopf, and V.~Vapnik.
\newblock Unifying distillation and privileged information.
\newblock {\em arXiv preprint arXiv:1511.03643}, 2015.

\bibitem{lopez2017gradient}
D.~Lopez-Paz and M.~A. Ranzato.
\newblock Gradient episodic memory for continual learning.
\newblock In {\em Advances in Neural Information Processing Systems}, pages
  6467--6476, 2017.

\bibitem{dqn}
V.~Mnih, K.~Kavukcuoglu, D.~Silver, A.~A. Rusu, J.~Veness, M.~G. Bellemare,
  A.~Graves, M.~Riedmiller, A.~K. Fidjeland, G.~Ostrovski, S.~Petersen,
  C.~Beattie, A.~Sadik, I.~Antonoglou, H.~King, D.~Kumaran, D.~Wierstra,
  S.~Legg, and D.~Hassabis.
\newblock Human-level control through deep reinforcement learning.
\newblock {\em Nature}, 518(7540):529--533, 2015.

\bibitem{moore1999multi}
A.~W. Moore, L.~Baird, and L.~P. Kaelbling.
\newblock Multi-value-functions: Effcient automatic action hierarchies for
  multiple goal {MDPs}.
\newblock In {\em Proceedings of the international joint conference on
  artificial intelligence}, pages 1316--1323, 1999.

\bibitem{oh2017zero}
J.~Oh, S.~Singh, H.~Lee, and P.~Kohli.
\newblock Zero-shot task generalization with multi-task deep reinforcement
  learning.
\newblock In {\em Proceedings of the 34th International Conference on Machine
  Learning-Volume 70}, pages 2661--2670. JMLR. org, 2017.

\bibitem{OpenAI_dota}
OpenAI.
\newblock Openai five.
\newblock \url{https://blog.openai.com/openai-five/}, 2018.

\bibitem{openai-shadowhand}
OpenAI, M.~Andrychowicz, B.~Baker, M.~Chociej, R.~J{\'{o}}zefowicz, B.~McGrew,
  J.~W. Pachocki, J.~Pachocki, A.~Petron, M.~Plappert, G.~Powell, A.~Ray,
  J.~Schneider, S.~Sidor, J.~Tobin, P.~Welinder, L.~Weng, and W.~Zaremba.
\newblock Learning dexterous in-hand manipulation.
\newblock {\em CoRR}, abs/1808.00177, 2018.

\bibitem{peters2005natural}
J.~Peters, S.~Vijayakumar, and S.~Schaal.
\newblock Natural actor-critic.
\newblock In {\em European Conference on Machine Learning}, pages 280--291.
  Springer, 2005.

\bibitem{riemer2018learning}
M.~Riemer, I.~Cases, R.~Ajemian, M.~Liu, I.~Rish, Y.~Tu, , and G.~Tesauro.
\newblock Learning to learn without forgetting by maximizing transfer and
  minimizing interference.
\newblock In {\em International Conference on Learning Representations}, 2019.

\bibitem{ring1994continual}
M.~B. Ring.
\newblock {\em Continual learning in reinforcement environments}.
\newblock PhD thesis, University of Texas at Austin Austin, Texas 78712, 1994.

\bibitem{ruder2017overview}
S.~Ruder.
\newblock An overview of multi-task learning in deep neural networks.
\newblock {\em arXiv preprint arXiv:1706.05098}, 2017.

\bibitem{rusu2015policy}
A.~A. Rusu, S.~G. Colmenarejo, C.~Gulcehre, G.~Desjardins, J.~Kirkpatrick,
  R.~Pascanu, V.~Mnih, K.~Kavukcuoglu, and R.~Hadsell.
\newblock Policy distillation.
\newblock {\em arXiv preprint arXiv:1511.06295}, 2015.

\bibitem{sareni1998fitness}
B.~Sareni and L.~Krahenbuhl.
\newblock Fitness sharing and niching methods revisited.
\newblock {\em IEEE transactions on Evolutionary Computation}, 2(3):97--106,
  1998.

\bibitem{saxe2013exact}
A.~M. Saxe, J.~L. McClelland, and S.~Ganguli.
\newblock Exact solutions to the nonlinear dynamics of learning in deep linear
  neural networks.
\newblock {\em ICLR}, 2014.

\bibitem{schaul2015prioritized}
T.~Schaul, J.~Quan, I.~Antonoglou, and D.~Silver.
\newblock Prioritized experience replay.
\newblock {\em arXiv preprint arXiv:1511.05952}, 2015.

\bibitem{barbados}
T.~Schaul, H.~van Hasselt, J.~Modayil, M.~White, A.~White, P.~Bacon, J.~Harb,
  S.~Mourad, M.~G. Bellemare, and D.~Precup.
\newblock The {Barbados} 2018 list of open issues in continual learning.
\newblock {\em CoRR}, abs/1811.07004, 2018.

\bibitem{shazeer2017outrageously}
N.~Shazeer, A.~Mirhoseini, K.~Maziarz, A.~Davis, Q.~Le, G.~Hinton, and J.~Dean.
\newblock Outrageously large neural networks: The sparsely-gated
  mixture-of-experts layer.
\newblock {\em arXiv preprint arXiv:1701.06538}, 2017.

\bibitem{alphago}
D.~Silver, A.~Huang, C.~J. Maddison, A.~Guez, L.~Sifre, G.~van~den Driessche,
  J.~Schrittwieser, I.~Antonoglou, V.~Panneershelvam, M.~Lanctot, S.~Dieleman,
  D.~Grewe, J.~Nham, N.~Kalchbrenner, I.~Sutskever, T.~Lillicrap, M.~Leach,
  K.~Kavukcuoglu, T.~Graepel, and D.~Hassabis.
\newblock Mastering the game of {Go} with deep neural networks and tree search.
\newblock {\em Nature}, 529:484--503, 2016.

\bibitem{alphazero}
D.~Silver, T.~Hubert, J.~Schrittwieser, I.~Antonoglou, M.~Lai, A.~Guez,
  M.~Lanctot, L.~Sifre, D.~Kumaran, T.~Graepel, et~al.
\newblock A general reinforcement learning algorithm that masters chess, shogi,
  and go through self-play.
\newblock {\em Science}, 362(6419):1140--1144, 2018.

\bibitem{silver2013lifelong}
D.~L. Silver, Q.~Yang, and L.~Li.
\newblock Lifelong machine learning systems: Beyond learning algorithms.
\newblock In {\em 2013 AAAI spring symposium series}, 2013.

\bibitem{stollenga2014deep}
M.~F. Stollenga, J.~Masci, F.~Gomez, and J.~Schmidhuber.
\newblock Deep networks with internal selective attention through feedback
  connections.
\newblock In {\em Advances in neural information processing systems}, pages
  3545--3553, 2014.

\bibitem{suttonbook}
R.~S. Sutton and A.~G. Barto.
\newblock {\em Reinforcement learning: An introduction}.
\newblock MIT press, 2018.

\bibitem{options}
R.~S. Sutton, D.~Precup, and S.~Singh.
\newblock Between mdps and semi-mdps: A framework for temporal abstraction in
  reinforcement learning.
\newblock {\em Artificial intelligence}, 112(1-2):181--211, 1999.

\bibitem{saxe-switching}
C.-Y. Tsai, A.~M. Saxe, and D.~Cox.
\newblock Tensor switching networks.
\newblock In D.~D. Lee, M.~Sugiyama, U.~V. Luxburg, I.~Guyon, and R.~Garnett,
  editors, {\em Advances in Neural Information Processing Systems 29}, pages
  2038--2046. Curran Associates, Inc., 2016.

\bibitem{alphastarblog}
O.~Vinyals, I.~Babuschkin, J.~Chung, M.~Mathieu, M.~Jaderberg, W.~M. Czarnecki,
  A.~Dudzik, A.~Huang, P.~Georgiev, R.~Powell, T.~Ewalds, D.~Horgan, M.~Kroiss,
  I.~Danihelka, J.~Agapiou, J.~Oh, V.~Dalibard, D.~Choi, L.~Sifre, Y.~Sulsky,
  S.~Vezhnevets, J.~Molloy, T.~Cai, D.~Budden, T.~Paine, C.~Gulcehre, Z.~Wang,
  T.~Pfaff, T.~Pohlen, Y.~Wu, D.~Yogatama, J.~Cohen, K.~McKinney, O.~Smith,
  T.~Schaul, T.~Lillicrap, C.~Apps, K.~Kavukcuoglu, D.~Hassabis, and D.~Silver.
\newblock {AlphaStar: Mastering the Real-Time Strategy Game StarCraft II}.
\newblock
  \url{https://deepmind.com/blog/alphastar-mastering-real-time-strategy-game-starcraft-ii/},
  2019.

\bibitem{reinforce}
R.~J. Williams.
\newblock Simple statistical gradient-following algorithms for connectionist
  reinforcement learning.
\newblock {\em Machine learning}, 8(3-4):229--256, 1992.

\bibitem{yin2017grad}
D.~Yin, A.~Pananjady, M.~Lam, D.~S. Papailiopoulos, K.~Ramchandran, and
  P.~Bartlett.
\newblock Gradient diversity empowers distributed learning.
\newblock {\em CoRR}, abs/1706.05699, 2017.

\end{thebibliography}

\section*{Acknowledgements}
We thank Hado van Hasselt, David Balduzzi, Andre Barreto, Claudia Clopath, Arthur Guez, Simon Osindero, Neil Rabinowitz, Junyoung Chung, David Silver, Remi Munos, Alhussein Fawzi, Jane Wang, Agnieszka Grabska-Barwinska, Dan Horgan, Matteo Hessel, Shakir Mohamed and the rest of the DeepMind team for helpful comments and suggestions.



\clearpage
\appendix
\onecolumn
\section{Additional results}
\begin{figure}[tbp]
\begin{center}
\vspace{-1em}
\centerline{
\includegraphics[width=0.5\columnwidth]{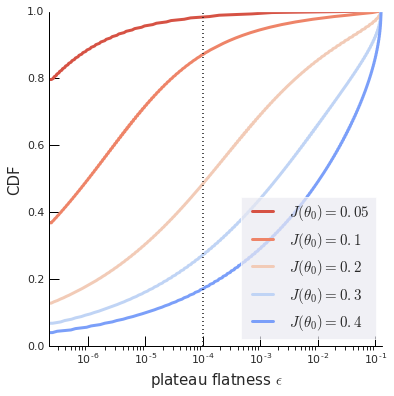}
\vspace{-1em}
}
\caption{Basins of attraction, for plateaus of different $\epsilon$, and for different levels of initial performance $J(\theta_0)$, under deterministic dynamics. The dashed line indicates the typical $\epsilon$ for which learning is 10 times slower than necessary (see \cref{fig:eps-badness}), so for example half of the trajectories initialized at $J(\theta_0)=0.2$ hit such a flat plateau.
\vspace{-1em}
}
\label{fig:eps-basins}
\end{center}
\end{figure}

We investigated numerous additional variants of the basic bandit setup.
In each case, we summarize the results by the probabilities that a plateau of $\epsilon$ or worse is encountered, as in \cref{fig:eps-basins}. We quantify this by computing the slowest progress along the learning curve (not near the start nor the optimum), normalized to factor out the step-size.
If not mentioned otherwise, we use the following settings across these experiments: $K=2$, $n=2$, low initial performance $J(\theta_0)=\frac{K}{10}$, step-size $\eta=0.1$, and batch-size $K$ (exactly 1 per context). Learning runs are stopped near the global optimum, when $J \geq K-0.1$, or after $N=10^5$ samples.

We can quantify the insights of \cref{sec:3dplus} by measuring the flatness of the \emph{worst} plateau in a learning curve that generally has more than one. \cref{fig:worst-plateau} gives results that validate the qualitative insights, when increasing $K$ and $n$ jointly.
Note that only scaling up $n$ actually makes the problem easier (when controlling for initial performance), because the actions $K < i \leq n$ are disadvantageous in all contexts, so there is some positive transfer through their action biases.

We looked at the influence of some architectural choices, using deep neural networks to parameterize the policy. It turns out that deeper or wider MLPs do not qualitatively change the dynamics from the simple setup in \cref{sec:bandit}.
\cref{fig:stepsizes} illustrates some of the effects of learning rates and optimizer choices.

\begin{figure}[tbp]
\begin{center}
\vspace{-1em}
\centerline{
\includegraphics[width=0.5\columnwidth]{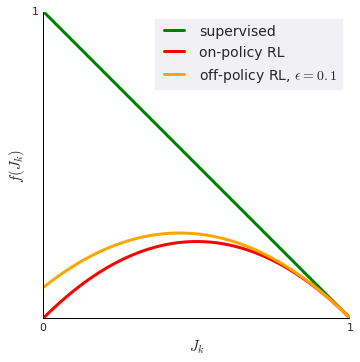}
\vspace{-1em}
}
\caption{Plot of the scalar coupling functions $f(J_k)$ of \cref{eq:ftimesv} (see \cref{sec:factored}). It highlights the U-shape for on-policy REINFORCE (red), in contrast to a supervised learning setup (green, see \cref{sec:supervised}). In orange, it illustrates how the $f(0)=0$ condition no longer holds when using off-policy data, in this case, mixing 90\% of on-policy data with 10\% of uniform random data.
\vspace{-1em}
}
\label{fig:wundt}
\end{center}
\end{figure}

\begin{figure}[tbp]
\begin{center}
\vspace{-1em}
\centerline{
\includegraphics[width=0.5\columnwidth]{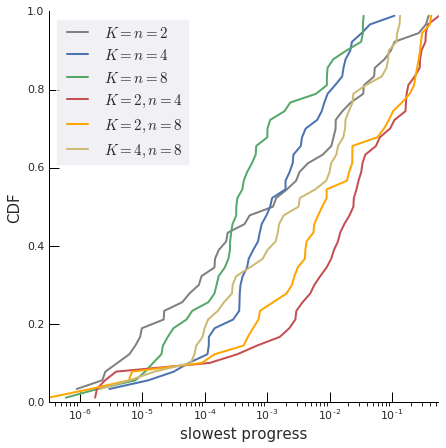}
\vspace{-1em}
}
\caption{Likelihood of encountering a plateau for different numbers of contexts $K$ and actions $n$ (same type of plot than \cref{fig:core-diffs}). Note how the positive transfer from spurious actions (when $K<n$, warm colors) helps performance.
\vspace{-1em}
}
\label{fig:worst-plateau}
\end{center}
\end{figure}

\begin{figure}[tbp]
\begin{center}
\vspace{-1em}
\centerline{
\includegraphics[width=0.5\columnwidth]{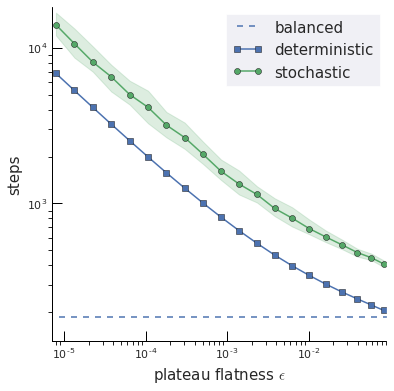}
\vspace{-1em}
}
\caption{Relation between $\epsilon$ of the traversed plateau, and the number of steps along a trajectory from near $(0,0)$ to near $(1,1)$.
The dashed line (`balanced') corresponds to trajectories that follow the diagonal ($J_1=J_2$) and don't encounter a plateau. Note that for $\epsilon\approx 10^{-4}$, the deterministic learning trajectories are 10 times slower than a diagonal trajectory (warm colors in \cref{fig:cartoon}).
\vspace{-1em}
}
\label{fig:eps-badness}
\end{center}
\end{figure}

\begin{figure*}[tbp]
\begin{center}
\vspace{-1em}
\centerline{
\includegraphics[width=0.5\columnwidth]{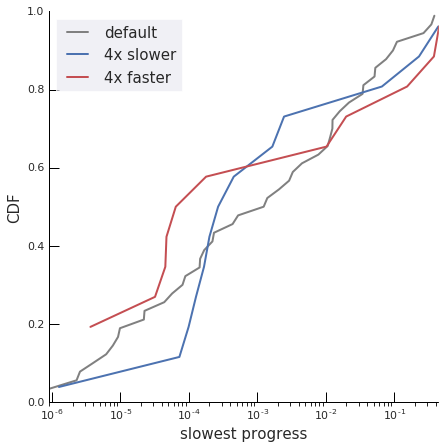}
\includegraphics[width=0.5\columnwidth]{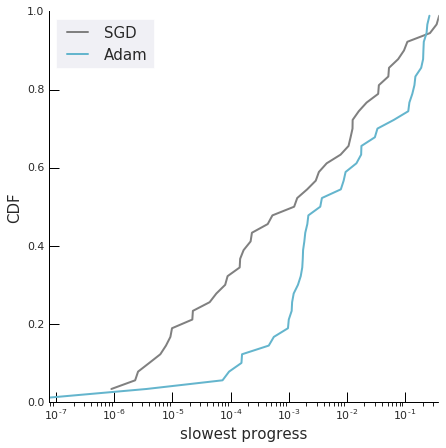}
\vspace{-1em}
}
\caption{
Likelihood of encountering a bad plateau (same type of plot than \cref{fig:core-diffs}). {\bf Left:} Comparison between step-sizes. {\bf Right:} Comparison between optimizers.
\vspace{-1em}
}
\label{fig:stepsizes}
\label{fig:optimizers}
\end{center}
\end{figure*}

\section{Detailed derivations (bandit)}

\subsection{Fixed point analysis}
\label{sec:fpa}

The Jacobian with respect to $J_1,J_2$ is
\[
\left[
\begin{array}{cc}
-2J_1+1 & J_2 - \frac{1}{2} \\
J_1- \frac{1}{2} & -2J_2+1
\end{array}
\right]
\]
with determinant $\frac{5}{4} (1-2J_1)(1-2J_2)$ and trace $-2J_1-2J_2+2$.
This lets us characterize the four fixed points:
\begin{center}
\begin{tabular}{c|cc|c}
fixed point & trace & determinant & type  \\
\hline
$0, 0$ & 2 & $\frac{5}{4}$ & unstable \\
$0, 1$ & 0 & $-\frac{5}{4}$ & saddle  \\
$1, 0$ & 0 & $-\frac{5}{4}$ & saddle  \\
$1, 1$ & -2 & $\frac{5}{4}$ & stable  \\
\end{tabular}
\end{center}

\subsection{Derivation of $\ddot{J}$ for RL}
\label{sec:ddot}

We characterize the \emph{acceleration} of the learning dynamics, as:
\begin{eqnarray*}
\ddot{J} & := &\lim_{\eta\rightarrow 0} \frac{\dot{J}\left(\theta + \eta\nabla J\right) - \dot{J}(\theta)}{\eta}
\\&=& \langle \nabla \dot{J}, \nabla J \rangle
\\&=& \langle \nabla [2J_1^2(1-J_1)^2 - 2J_1(1-J_1) J_2(1-J_2) + 2J_2^2(1-J_2)^2 ], \nabla J \rangle
\\&=& \langle 4J_1(1-J_1)(1-2J_1)\nabla J_1 + 4J_2(1-J_2)(1-2J_2) \nabla J_2 
\\&&
- 2J_1(1-J_1)(1-2J_2) \nabla J_2- 2J_2(1-J_2)(1-2J_1) \nabla J_1, \nabla J_1 + \nabla J_2 \rangle
\\&=& 2(1-2J_1) [2J_1(1-J_1) - J_2(1-J_2)] \|\nabla J_1\|^2
+ 2(1-2J_2) [-J_1(1-J_1) + 2J_2(1-J_2)] \|\nabla J_2\|^2
\\&&
+ 2[(1-2J_1) [2J_1(1-J_1) - J_2(1-J_2)] 
+ (1-2J_2) [-J_1(1-J_1) + 2J_2(1-J-2)]]\langle \nabla J_1, \nabla J_2 \rangle
\\&=& 4(1-2J_1)(2J_1-J_2)J_1^2 + 4(1-2J_2)(2J_2-J_1)J_1^2
-2J_1J_2[(1-2J_1)(2J_1-J_2) + (1-2J_2)(2J_2-J_1)]
\\&=& 2(1-J_1-J_2)\left[-8 J_1^6 + 8 J_1^5 J_2 + 20 J_1^5 - 20 J_1^4 J_2 - 16 J_1^4 \right.
- 2 J_1^3 J_2^3 + 3 J_1^3 J_2^2 + 15 J_1^3 J_2 + 4 J_1^3 + 3 J_1^2 J_2^3 
\\&& \hspace{7em}\left. - 4 J_1^2 J_2^2 - 3 J_1^2 J_2 + 8 J_1 J_2^5 - 20 J_1 J_2^4 + 15 J_1 J_2^3 
- 3 J_1 J_2^2 - 8 J_2^6 + 20 J_2^5 - 16 J_2^4 + 4 J_2^3\right]
\\&:=&(1-J_1-J_2)P_6(J_1,J_2)
\enspace,
\end{eqnarray*}
This implies that $\ddot{J}=0$ along the diagonal where $J_2=1-J_1$, and $\ddot{J}$ changes sign there. We have a plateau if the sign-change is from negative to positive, 
in other words, wherever 
\begin{eqnarray*}
P_6(J_1, 1-J_1) \leq 0
&\Leftrightarrow & -(J_1 - 1)^2 J_1^2 (30 J_1^2 - 30 J_1 + 7) \leq 0
\\
&\Leftrightarrow &J_1 \in \left[0, \frac{15-\sqrt{15}}{30}\right] \operatorname{or} J_1 \in \left[\frac{15+\sqrt{15}}{30}, 1\right]
\enspace,
\end{eqnarray*}
see also \cref{fig:clines}.

\subsection{Lower bound on basin of attraction}
\label{sec:lower-bound}
We can construct an explicit lower bound on the size of the basing of attraction for a given plateau.
The main argument is that once a trajectory is in a WTA region (say, the one with $\dot{J_1}<0$), it can only leave it after the dominant component is nearly learned ($J_2\approx 1$), and the performance on the dominated component $J_1$ has regressed.
$(J_{1_{[0]}},J_{2_{[0]}})\sim P_0$, where
Note that we abuse notation, 
$J_{1_{[0]}} = J_1(\theta_0)$, and $P_0$ technically describes the distribution of $\theta_0$.
Note that trajectories that leave WTA region by crossing by crossing the null cline $\dot{J_1}=0$ or $\dot{J_2}=0$ traverses diagonal at an $\epsilon$-plateau point. If we are to consider traditional initialization of the neural network we can look at the probability of the initialization to be underneath either null cline.
Under assumption on the initialization (assuming uniform distribution across angular directions) we have over $50\%$ chance of initializing the model in a WTA region. 
For this we can compute the derivative of the null cline equation at the origin, 
and assuming a distribution that is uniform across angular directions we have approximately $59\%$ chance to start in a WTA region. 
Let us consider the dynamics around the top left saddle point. A trajectory that leaves the WTA region by crossing the $\dot{J_1}=0$ null cline at $(J_{1_{[1]}}, J_{2_{[1]}})$ will traverse the diagonal at an $\epsilon$-plateau point $(J_{1_{[2]}}, 1-J_{1_{[2]}})$ where $\epsilon \approx 2J_{1_{[2]}}^2$ and $J_{1_{[2]}} \leq \frac{1}{2}(1+J_{1_{[1]}}-J_{2_{[1]}})$, because $\dot{J_1} \leq \dot{J_2}$ in that region.
Furthermore, for any trajectory that starts at a point $(J_{1_{[0]}}, J_{2_{[0]}})$ within the WTA region to the left of this null cline, if $J_{1_{[0]}} \leq J_{1_{[1]}}$, then it will exit the WTA region at $(J_{1_{[1]}}, J_{2_{[1]}})$ or above, and thus hit a plateau that is at least as flat.
So the basin of attraction for an $\epsilon$-plateau includes the polygon defined by $(0,0), (0,1), (J_{1_{[2]}},1-J_{1_{[2]}}), (J_{1_{[1]}}, J_{2_{[1]}}), (J_{1_{[1]}}, 1-J_{2_{[1]}})$, but is in fact larger, as other trajectories can enter this region from elsewhere, see \cref{fig:clines}.
for a diagram with this geometric intuition, and \cref{fig:eps-badness} for the empirical relation between $\epsilon$ and basin size.
In these simulations, and elsewhere if not mentioned otherwise, w
We use $\eta=0.1$ to produce trajectories, and exactly one sample per context for stochastic updates.

\subsection{Probability WTA initialization near origin}
\label{sec:init-prob}
When the distribution of starting points is close to the origin, then a quantity of interest is the probability $P_{\text{nc}}$ of a starting point falling underneath either null cline (because from there on the WTA dynamics will pull it into a plateau). For this we can compute the derivatives of the null cline equation at the origin:
\begin{eqnarray*}
\left.\frac{d}{dJ_1} [2J_1(1-J_1)]\right|_{J_1=0} = 2
&&
\left.\frac{d}{dJ_2} [J_2(1-J_2)]\right|_{J_2=0} = 1
\end{eqnarray*}
So for such a distribution that is uniform across angular directions we have $P_{\text{nc}} = \frac{4}{\pi}\arctan\left(\frac{1}{2}\right)\approx 59\%$.
However, as \cref{fig:eps-basins} shows empirically, the basins of attraction of flat plateaus are even larger, because starting in a WTA region is sufficient but not necessary.

\subsection{Derivation of $\ddot{J}$ for supervised learning}
\label{sec:sup-detail}
We have:
\begin{eqnarray*}
J_{\text{sup}}  &=& \log J_1 + \log J_2
\\ \nabla J_1 &=& J_1(1-J_1) \dxvec
\\ \nabla J_2 &=& J_2(1-J_2) \dyvec
\\  \nabla J_{\text{sup}} &=&   
\nabla  \log J_1 + \nabla  \log J_2 = \frac{\nabla J_1}{J_1} + \frac{\nabla J_2}{J_2}
=(1-J_1) \dxvec + (1-J_2) \dyvec 
\\ \rho_{\text{sup}} &=& \frac{\langle \nabla \log J_1, \nabla \log J_2 \rangle}
{\|\nabla J_1\| \|\nabla J_2\|}
= - \frac{1}{2}
\\
\dot{J_1}  &=& \langle \nabla J_1, \nabla J \rangle
=2J_1(1-J_1)^2 - J_1(1-J_1)(1-J_2)
=J_1(1-J_1)(1-2J_1+J_2)
\\ \dot{\log J_1} &=& \frac{\dot{J_1}}{J_1} = 2(1-J_1)^2 - (1-J_1)(1-J_2)
\\ \dot{J}_{\text{sup}} &=&  \dot{\log J_1} + \dot{\log J_2} 
=2(1-J_1)^2 - 2(1-J_1)(1-J_2) + 2(1-J_2)^2 
=2(J_1-J_2)^2 + 2(1-J_1)(1-J_2)
\\&\geq&0
\\
 \nabla \dot{J}_{\text{sup}} &=& -4(1-J_1)\nabla J_1 -4(1-J_2)\nabla J_2 
 +2(1-J_1)\nabla J_2 + 2(1-J_2)\nabla J_1
\\ &=& (4J_1 - 2J_2 -2)\nabla J_1 + (4J_2 - 2J_1 -2)\nabla J_2
\\ \ddot{J}_{\text{sup}} &=& \langle \nabla \dot{J}_{\text{sup}}, \nabla J_{\text{sup}} \rangle
\\&=& (4J_1 - 2J_2 -2)\dot{J_1} + (4J_2 - 2J_1 -2)\dot{J_2}
\\&=& - 2J_1(1-J_1)(1-2J_1 +J_2 )^2 - 2J_2(1-J_2)(1-2J_2 +J_1)^2
\\ &\leq& 0
\end{eqnarray*}

\section{Derivations for factored objectives case}

\subsection{Derivation of $\ddot{J}$}
\label{sec:ddot-gen}
We consider objectives where each component is smooth, and can be written in the following form:
\begin{eqnarray*}
\nabla_\theta J_k(\theta) &=& f_{k}(J_k) v_k(\theta)
\end{eqnarray*}
where $f_k : \R \mapsto \R^+$ is a scalar function mapping to positive numbers that doesn't depend on current $\theta$ except via $J_k$ and $v_k : \R^d \mapsto \R^d$.
We further assume that each component is bounded: $0 \leq J_k \leq J_k^{\max}$. When the optimum is reached, there is no further learning, so $f_k(J_k^{\max}) = 0$, but for intermediate points learning is always possible, i.e., $0 < J_k < J_k^{\max} \Rightarrow  f_k(J_k) >0$.

If the above conditions hold, then these are sufficient conditions to show that the combined objective $J$ admits an $\epsilon$-plateau as defined in Definition 2. Moreover, $J$ will exhibit the saddle points at $J=(0,J_{max})$ and $J=(J_{max},0)$.

\begin{eqnarray*}
\rho   &=& \frac{\langle \nabla J_1, \nabla J_2 \rangle}{\| \nabla J_1\| \| \nabla J_2\|} 
= \frac{v_1^{\top}v_2}{\|v_1\| \|v_2\|}
\\ \dot{J_1} &=& \langle \nabla J_1, \nabla J \rangle = \|v_1\|^2 f_1(J_1)^2 + \rho \|v_1\| \|v_2\| f_1(J_1) f_2(J_2)
\\&=& \|v_1\| f_1(J_1) \left[\|v_1\| f_1(J_1) + \rho \|v_2\|  f_2(J_2)\right]
\\\dot{J}&=& \langle \nabla J, \nabla J \rangle = \langle \nabla J_1 + \nabla J_2,  \nabla J_1 + \nabla J_2 \rangle
\\&=&\|v_1\|^2 f_1(J_1)^2  +\|v_2\|^2 f_2(J_2)^2  + 2\rho \|v_1\| \|v_2\| f_1(J_1) f_2(J_2)
\\&=& v_1^{\top}v_1 f_1(J_1)^2  + v_2^{\top}v_2f_2(J_2)^2  +  2f_1(J_1) f_2(J_2) v_1^{\top}v_2
%
\\
\nabla \dot{J}& = &
2\|v_1\|^2 f'_1(J_1)f_1(J_1)\nabla J_1
+2\|v_2\|^2 f'_2(J_2)f_2(J_2)\nabla J_2
+ 2\rho \|v_1\| \|v_2\| f'_1(J_1)f_2(J_2)\nabla J_1
\\&&
+ 2\rho \|v_1\| \|v_2\| f'_2(J_2)f_1(J_1)\nabla J_2
+2 f_1(J_1)^2 v_1^{\top}\nabla v_1 
+2 f_2(J_2)^2 v_2^{\top}\nabla v_2
\\&&
+ 2f_1(J_1) f_2(J_2) v_2^{\top}\nabla v_1
+ 2f_1(J_1) f_2(J_2) v_1^{\top}\nabla v_2
\end{eqnarray*}
For the special case $\ddot{J}|_{J_1=0}$ where $f_1(J_1)=f_1(0)=0$, the majority of terms in the expression vanish:
\begin{eqnarray*}
\left.\nabla \dot{J}\right|_{J_1=0} &=& 2\|v_2\|^2 f'_2(J_2)f_2(J_2)\nabla J_2 + 2 f_2(J_2)^2 v_2^{\top}\nabla v_2
\end{eqnarray*}
and
\begin{eqnarray*}
\left.\ddot{J}\right|_{J_1=0} 
&=& \langle \nabla \dot{J}, \nabla J_1 + \nabla J_2 \rangle 
\\ &=&\langle 2\|v_2\|^2 f'_2(J_2)f_2(J_2)\nabla J_2 + 2 f_2(J_2)^2 v_2^{\top}\nabla v_2, \nabla y \rangle
\\ &=& 2\|v_2\|^4 f'_2(J_2)f_2(J_2)^3 + 2 f_2(J_2)^3 v_2^{\top}\nabla v_2 v_2
\end{eqnarray*}
If we assume that the $\nabla v_k = \frac{dv_k}{d\theta}$ terms are negligibly small near the saddle points, then we obtain
\begin{eqnarray*}
\left.\ddot{J}\right|_{J_1=0} &\approx& 2\|v_2\|^4 f'_2(J_2)f_2(J_2)^3
\end{eqnarray*}
By symmetry, the result for $\ddot{J}|_{J_2=1}$ is very similar. 

\subsection{Connecting the WTA region and the plateau}
\label{sec:plateau-wta}

Further, when ignoring the $\nabla v_k = \frac{dv_k}{d\theta}$ terms, we get the general from
\begin{eqnarray*}
\nabla \dot{J}
&=& 
2\|v_1\|(\|v_1\|f_1(J_1) + \rho\|v_2\| f_2(J_2)) f'_1(J_1) \nabla J_1 
+2\|v_2\|(\|v_2\|f_2(J_2) + \rho\|v_1\| f_1(J_1)) f'_2(J_2) \nabla J_2
\\&=&
2\frac{\dot{J_1}}{f_1(J_1)}f'_1(J_1) \nabla J_1 + 2\frac{\dot{J_2}}{f_2(J_2)}f'_2(J_2) \nabla J_2
\\ \ddot{J} &=& 2\frac{\dot{J_1}}{f_1(J_1)}f'_1(J_1) \|\nabla J_1\|^2 + 2\frac{\dot{J_2}}{f_2(J_2)}f'_2(J_2) \|\nabla J_2\|^2 
+2\left[\frac{\dot{J_1}}{f_1(J_1)}f'_1(J_1)  + \frac{\dot{J_2}}{f_2(J_2)}f'_2(J_2)\right]\nabla J_1^{\top} \nabla J_2
\\&=&2 \|v_1\|^2 \dot{J_1} f_1(J_1) f'_1(J_1) +2 \|v_2\|^2 \dot{J_2} f_2(J_2) f'_2(J_2) 
+2 \rho \|v_1\| \|v_2\|[\dot{J_1}f_2(J_2)f'_1(J_1) + \dot{J_2}f_1f'_2(J_2)]
\\&=&2\dot{J_1}f'_1(J_1)\|v_1\|(\|v_1\|f_1(J_1)+\rho\|v_2\|f_2(J_2)) + \ldots
\\&=& 2\frac{f'_1(J_1)}{f_1(J_1)} (\dot{J_1})^2 + 2\frac{f'_2(J_2)}{f_2(J_2)} (\dot{J_2})^2
\end{eqnarray*}
From this we can deduce that the sign change in $\ddot{J}$ must occur in the region between the null clines $\dot{J_1}=0$ and $\dot{J_2}=0$.

\end{document}